%% file: main.tex
\documentclass[manuscript]{acmart} 

\usepackage{balance}
\AtBeginDocument{%
  \providecommand\BibTeX{{%
    \normalfont B\kern-0.5em{\scshape i\kern-0.25em b}\kern-0.8em\TeX}}}

\copyrightyear{2023}
\acmYear{2023}
\setcopyright{rightsretained}
\begin{document}

\title[Detecting LLM-Generated Text in Computing Education]{Detecting LLM-Generated Text in Computing Education: A Comparative Study for ChatGPT Cases}


% according to: https://tex.stackexchange.com/questions/424066/acmart-multiple-authors-all-with-same-affiliation-one-author-an-additional-af
% this is due to long email issue of Michael
\author{Michael Sheinman Orenstrakh}
\email{michael.sheinmanorenstrakh@mail.utoronto.ca}
\affiliation{
  \institution{University of Toronto Mississauga}
  \city{Mississauga}
  \country{Canada}
}

\author{Oscar Karnalim}
\email{oscar.karnalim@it.maranatha.edu}
\orcid{0000-0003-4930-6249}
\affiliation{
  \institution{Maranatha Christian University}
  \city{Bandung}
  \country{Indonesia}
}

\author{Carlos Aníbal Suárez}
\email{carasuar@espol.edu.ec}
\orcid{0000-0002-6012-932X}
\affiliation{
  \institution{Escuela Superior Politécnica del Litoral}
  \city{Guayaquil}
  \country{Ecuador}
}

\author{Michael Liut}
\email{michael.liut@utoronto.ca}
\orcid{0000-0003-2965-5302}
\affiliation{
  \institution{University of Toronto Mississauga}
  \city{Mississauga}
  \country{Canada}
}

% \author{Michael Sheinman Orenstrakh}
% \affiliation{
%   \institution{University of Toronto Mississauga}
%   \country{Canada}
% }
% \email{michael.sheinmanorenstrakh@mail.utoronto.ca}

% \author{Michael Liut}
% \affiliation{
%   \institution{University of Toronto Mississauga}
%   \country{Canada}
% }
% \email{michael.liut@utoronto.ca}

% \author{Oscar Karnalim}
% \affiliation{%
%   \institution{Maranatha Christian University}
%   \country{Indonesia}
% }
% \email{oscar.karnalim@it.maranatha.edu}


\begin{abstract}
% preliminary abstract..  
% Motivation
% Objectives/Goals
% Methods
% Results
Due to the recent improvements and wide availability of Large Language Models (LLMs), they have posed a serious threat to academic integrity in education. Modern LLM-generated text detectors attempt to combat the problem by offering educators with services to assess whether some text is LLM-generated. In this work, we have collected $124$ submissions from computer science students before the creation of ChatGPT. We then generated $40$ ChatGPT submissions. We used this data to evaluate eight publicly-available LLM-generated text detectors through the measures of accuracy, false positives, and resilience. The purpose of this work is to inform the community of what LLM-generated text detectors work and which do not, but also to provide insights for educators to better maintain academic integrity in their courses. Our results find that CopyLeaks is the most accurate LLM-generated text detector, GPTKit is the best LLM-generated text detector to reduce false positives, and GLTR is the most resilient LLM-generated text detector. We also express concerns over $52$ false positives (of $114$ human written submissions) generated by GPTZero. Finally, we note that all LLM-generated text detectors are less accurate with code, other languages (aside from English), and after the use of paraphrasing tools (like QuillBot). Modern detectors are still in need of improvements so that they can offer a full-proof solution to help maintain academic integrity. Further, their usability can be improved by facilitating a smooth API integration, providing clear documentation of their features and the understandability of their model(s), and supporting more commonly used languages.
\end{abstract}




\maketitle

\input{body.tex}

% \section*{Acknowledgment}
% It has been removed for the dual-anonymous review process.

% \begin{acks}

% \end{acks}

\balance{}

\bibliographystyle{ACM-Reference-Format}
\bibliography{main.bib}

\end{document}

%% file: body.tex
\section{Introduction} 
In academia, a way to encourage students utilizing all learning opportunities and experiences is to properly maintain academic integrity in the courses \cite{Lancaster2018Academic}. Students need to complete any exams and assessments with their best effort. Further, they need to actively engage with the instructors (and tutors).

Although Artificial Intelligence (AI) can foster education \cite{Chen2020Artificial}, it might be misused to breach academic integrity. Paraphrasing tools \cite{Prentice2018Paraphrasing} and code obfuscation tools \cite{behera2015different} for example, are misused to cover up evidence for plagiarism (a breach of academic integrity about copying one's work and reusing it without proper acknowledgment \cite{Fraser2014Collaboration}).

Misuse of AI chatbots with large language models (LLM) \cite{carlini2021extracting} such as ChatGPT\footnote{\url{https://openai.com/blog/chatgpt}} is another trending threat for breaching academic integrity. Students can complete exams or assessments with limited effort, resulting in questionable performance; it is unclear whether the learning objectives are actually met. The misuse can be considered as contract cheating (i.e., getting help in exchange for mutual incentives \cite{lancaster2016contract}) since AI chatbots provide responses in exchange for additional user data. However, considering AI responses are generated based on other people's textual data without proper acknowledgment, we believe it is more justifiable to consider the misuse as plagiarism.

While checking student work for plagiarism, instructors are often aided by automated detectors. A number of detectors have been developed to detect whether a work is a result of LLM. Two of them are GPT-2 Output Detector  \cite{Solaiman2019Release} and Giant Language model Test Room (GLTR) \cite{Sebastian2019GLTR}. Nevertheless, due to the recency of misuse of AI chatbots, Computing educators might have limited information about publicly available detection detectors. Further, it is challenging to choose the most suitable detector for their teaching environment. To the best of our knowledge, there are no empirical studies comparing the detectors in terms of effectiveness.

In response to the aforementioned gaps, we investigate LLM-generated text detectors and formulate the following research question (RQ): ``How effective are LLM-generated text detectors?''

It is clear that there is a need in the community to understand if the currently available detectors are able to detect LLM-generated content~\cite{tang2023science, sadasivan2023aigenerated, perkins2023game} and what there reliability is.

As an additional contribution, we also report our experience in using the LLM-generated text detectors. It might be useful for readers interested in employing those detectors in their classrooms.

\section{Related Work}
This section discusses common breaches of academic integrity in computing education and misuse of AI to breach academic integrity.

\subsection{Common Breaches of Academic Integrity}

Academic integrity encourages students to act honestly, trustworthy, respectfully, and responsibly in learning\footnote{\url{https://lo.unisa.edu.au/course/view.php?id=6751\&amp;section=6}}. \citet{Lancaster2018Academic} lists five common breaches of academic integrity in computing education: plagiarism, collusion, contract cheating, exam cheating, and research fraud. It is important to inform students about instructors' expectations about academic integrity in their courses \cite{Simon2018Informing} and penalize those who breach academic integrity. 

Plagiarism happens when ideas, words, or even code is reused without proper acknowledgment and permission to the original author(s) \cite{Fraser2014Collaboration}.
It is commonly identified with the help of automated detectors \cite{Blanchard2022Stop} such as Turnitin\footnote{\url{https://www.turnitin.com/}}, Lichen \cite{Peveler2019Lichen}, MOSS\footnote{\url{https://theory.stanford.edu/~aiken/moss/}}, and JPlag \cite{Prechelt2002Finding}. Any submissions with high similarity will be investigated and if they are indeed a result of misconduct, the students will be penalized \cite{Karnalim2019Similarity}. 

Nevertheless, identifying plagiarism is not always straightforward; some perpetrators disguise their act with automated paraphrasing \cite{Prentice2018Paraphrasing, krishna2023paraphrasing}, essay spinning \cite{lancaster2009automated} or code obfuscation \cite{behera2015different}. The automated detectors should be resilient to common disguising practices in addition to being effective and efficient.
GPlag \cite{Liu2006Gplag} and BPlag \cite{Cheers2021Academic} for examples, focus on content semantic while measuring similarity among submissions. 
\citet{Tahaei2018Automated} and \citet{Yan2018TMOSS} developed detectors that detect substantial changes among consecutive saves. 
\citet{Ljubovic2020Plagiarism} developed a detector that is automatically integrated to a programming workspace to record any code edits.

Collusion is also about reusing ideas, words, or code without proper acknowledgment. However, the original author(s) is aware about the matter and somewhat allows it \cite{Fraser2014Collaboration}. Typically, this occurs when two or more students work closely beyond reasonable levels of collaboration \cite{Lancaster2018Academic}. Collusion can be identified in the same manner as plagiarism with the help of automated detectors. Similar submissions are reported by the detectors and then manually investigated by the instructors; students whose submissions are indeed a result of misconduct will be penalized.

Contract cheating occurs when third parties are paid to complete student assessments \cite{lancaster2016contract}. The third parties can be professional companies or even their colleagues. Contract cheating is quite challenging to identify as the third parties tend to know how to evade detection. It is only identifiable when the writing style and the quality of the submission is substantially different to those of the student's prior submissions. To expedite the identification process, instructors can either use the help of authorship identification detectors \cite{kale2017systematic} such as Turnitin Authorship Investigate\footnote{\url{https://help.turnitin.com/MicroContent/authorship-investigate.htm}} \cite{Dawson2020Software} or check contract cheating sites \cite{Rowland2018Turn}.

Exam cheating happens when some students have unfair advantages in the exams \cite{Lancaster2018Academic}. The advantages can vary from concealed notes during exams, leaked exam questions, to impersonation (i.e., an individual switch places with a student to take the exam). Exam cheating can be identified via careful investigation on the whole process of the exams. Sometimes, such identification can be aided with online proctoring systems \cite{Dendir2020Cheating} (e.g., Proctorio\footnote{\url{https://proctorio.com/}} and ProctorExam\footnote{\url{https://proctorexam.com/}}) or local monitoring tools (e.g., NetSupport\footnote{\url{https://www.netsupportschool.com/}}).

Research fraud means reporting research results without verifiable evidence \cite{Lancaster2018Academic}. It can be data fabrication (i.e., generating artificial data to benefit the students) or data falsification (i.e., updating the data so that it aligns with the students' desired findings). Both are parts of research misconduct\footnote{\url{https://grants.nih.gov/policy/research_integrity/definitions.htm}} and they can happen in research-related assessments. Research fraud can be identified via careful investigation on the whole process of research. Due to its complex nature, such misbehaviour is manually identified on most cases. However, instructors can get some help from source metadata \cite{Yamamoto2018Understanding} and automated image manipulation detection \cite{Bucci2017Automatic}.

\subsection{Misuse of AI}

AI substantially affects education \cite{Chen2020Artificial}. It improves student learning experience via the help of intelligent tutoring systems \cite{Elham2021Intelligent} and personalized learning materials \cite{Li2021Features}. AI expedites the process of providing feedback \cite{Cavalcanti2021Automatic}, identifying breaches of academic integrity \cite{Ullah2018Software}, maintaining student retention \cite{Albreiki2021Systematic}, learning programming \cite{puryear2022github}, creating programming exercises \cite{ICER22-Denny}, and recording attendance \cite{Budi2018Ibats}.

Advances in AI might also be misused for breaching academic integrity.
Paraphrasing tools \cite{Prentice2018Paraphrasing} which are intended to help students learn paraphrasing are misused to cover up plagiarism. 
Code generators like GitHub Copilot \cite{githubcopilot2023} which are intended to help programmers in developing software are misused to complete programming tasks that should be solved independently.
Code obfuscation tools \cite{behera2015different} which are intended to secure code in production are misused to disguise similarities in copied code submissions.

AI chatbots \cite{Nicolescu2022Human}, especially those with Large Language Model (LLM) \cite{carlini2021extracting} are intended to help people searching information, but they are misused to unethically complete exams\footnote{\url{https://edition.cnn.com/2023/01/26/tech/chatgpt-passes-exams/index.html}} and assessments\footnote{\url{https://theconversation.com/chatgpt-students-could-use-ai-to-cheat-but-its-a-chance-to-rethink-assessment-altogether-198019}}. 
LLM is derived from Language Model (LM), a statistical model at which each sequence of words are assigned with a probability \cite{Croft2010Search}. Per query or question, the response is generated by concatenating sequences of words that have high probability with the query or the question.

ChatGPT is a popular example of LLM. The tool is developed by OpenAI, a non-profit American research laboratory on top of GPT-3, a LLM with deep learning to generate human-like text. The tool relies on reinforcement and supervised learning to further tune the model.

A number of automated detectors have been developed to help instructors identifying AI misuses for breaching academic integrity. In the context of plagiarism and collusion, automated detectors nullify common alterations that can be done without understanding the content \cite{Ragkhitwetsagul2018Comparison, kvrivzkova2016preference} and remove contents that are not evident for raising suspicion \cite{Simon2020Choosing}.
In dealing with misuses of AI chatbots, a few automated detectors are developed under the same way as the chatbots via pretrained model, but dedicated to detect AI-generated texts. GPT-2 Output Detector  \cite{Solaiman2019Release} and GLTR \cite{Sebastian2019GLTR} are two of the examples.

\section{Methodology}
This section discusses how the research question stated in the introduction would be addressed and our preliminary work to discover publicly available LLM-generated text detectors.

We collected historical assignment data dating back to 2016 from two publicly funded research-focused institutions, one in North America and one in South America. The data collected was from upper-year undergraduate computer science and engineering students. 

We analyzed a total of $164$ submissions ($124$ were submitted by humans, $30$ were generated using ChatGPT, and $10$ were generated by ChatGPT and altered using the Quillbot paraphrasing tool) and compared them against eight LLM-generated text detectors. This results in a total of $1,312$ prediction results. 

Of the $164$ submissions, $134$ were written in English ($20$ of which were generated by a LLM, and another $10$ which were LLM-generated and paraphrased) and $20$ were written in Spanish ($10$ of which were AI-generated). The submissions were collected between $2016$ and $2018$ (prior to the release of ChatGPT), and were made in ``databases'', ``networking'', and a ``final thesis project'' course. These courses were specifically selected as they are upper-year computer science major courses that touch on a mix of systems and theory (databases and networking), as well as technical writing in computer science with a programming/development component (final thesis project). The students in these courses were primarily in a computer science major. It should also be noted that Spanish was selected as an alternative language to analyse because it is one of the world's most popular languages, and some of the authors have experience writing and evaluating technical material in this language.

The assessments analyzed in this study (see Table~\ref{tab:questions}) are taken from three undergrad courses. The first course is a databases course offered to third-year computer science students in their first or second semester. It is a mix of database theory and practical systems application. There are $101$ paper submissions from this course which involved a final assessment where students wrote a report analyzing two industry players and their use of databases and data centers, this was written in English. 

The second course is a networking course offered to third-year computer science students in their second semester. It is a mix of theoretical concepts and practical system application. There are $13$ paper submissions from this course which involved an exam question where students explain how they would implement the NOVEL-SMTP and NEO-SMTP email protocols using only UDP, this was written in English. 

The third course is a final thesis project course offered to fourth-year computer science students throughout their final year of study (across both semesters). It is meant to bridge theory and practice to develop something that can be used/implemented in the real world. There are $10$ paper submissions from this course which involved improving computing systems and engineering processes in their local community, this was written in Spanish.

Due to the character limitations, data below 1,000 characters was excluded and data above 2,500 characters was truncated to the last complete sentence. This ensures the input data fits within the range of all detectors. As many LLM-generated text detection platforms have a 2,500 character maximum, to ensure fairness across platform, we used 2,500 characters as our upper-bound.

LLM-generated texts were created with the help of ChatGPT\footnote{\url{https://openai.com/blog/chatgpt}}, a popular LLM. The handouts were parsed to prompts by removing irrelevant information (course code, deadlines, submission instruction) so the prompts only contain the core requirements of the task. These prompts were then fed into ChatGPT to generate a solution to the assignment.  

It should be noted, the authors mined through over $2,000$ submissions in programming, data structures and algorithms, and compilers courses, however, the submission data varied too much for the content to easily be extracted and analyzed for detectors. Often due to a lack of context after removing any code. The selected submissions were purely writing-based and did not involve coding components, they did in some cases discuss theoretical concepts in computer science. 

Finally, all of the detectors were tested in April 2023.

% data sets: ChatGPT and student submissions

% TODO: I did this, although we are not reporting any results from here. The data was highly latex and math based so it was hard to make definite correlations. 
% Taking this out for now
% As an additional data set, we also collected student submissions after the release of ChatGPT (2022) and check their originality. Any suspected submissions (with less than 50\% originality) would be manually investigated to determine whether they are actually a result of using ChatGPT. \\

\begin{table}
\caption{Questions analyzed from submissions}
\label{tab:questions}
\begin{tabular}{|p{2cm}|p{6cm}|}
\hline
\textbf{Course} & \textbf{Question}             \\ \hline
Databases             & Write a report analyzing two industry players and their use of databases and data centers. Discuss the efficiency, scalability, and social impact. \\\hline
Networking            & Implement the NOVEL-SMTP and NEO-SMTP email protocols using only UDP at the transport layer.  \\\hline
Thesis Project              & Continuous Improvement and Operational Excellence in Manufacturing and Industrial Processes. \\\hline
\end{tabular}
\end{table}

\subsection{Discovering Publicly Available LLM-generated Text Detectors}
Publicly available LLM-generated text detectors were discovered from January to February 2023 from social media (i.e., Twitter, Facebook, and blogs), online news, and previous literature on LLM-generated text detection (GPT-2, GLTR). Public interest in LLM-generated text detectors followed the release of GPTZero which went viral on January, 2023. After GPTZero, many other companies launched their own LLM-generated text detectors.  

A number of LLM-generated text detectors were discovered but we limited this study to LLM-generated text detectors that appear to offer proprietary solutions to LLM-generated text detection. We found that some LLM-generated text detectors are likely to be replicas of open source work (GPT-2) and hence we excluded such detectors from the study. 

We identified eight such publicly available LLM-generated text detectors, as shown in Table \ref{tab:detectors}. Two of them (GPT-2 Output Detector and GLTR) are featured with technical reports~\cite{Solaiman2019Release,Sebastian2019GLTR}.\\

\begin{table}
  \caption{Discovered publicly available LLM-generated text detectors; model info refers to how detailed the information of the used LLM (complete, partial, and none)}
  % . M.U. = Model Understandability.}
  \label{tab:detectors}
  \begin{tabular}{p{2cm} p{4cm} p{1.5cm}}
    \toprule Name & Link & Model Info \\ \midrule
    GPT-2 Output Detector & \url{https://openai-openai-detector.hf.space/} & Complete \\
    GLTR  & \url{http://gltr.io/dist/index.html} & Complete \\
    CopyLeaks & \url{https://copyleaks.com/features/ai-content-detector} & None \\
    GPTZero & \url{https://gptzero.me/} & Partial \\
    AI Text Classifier & \url{https://platform.openai.com/ai-text-classifier} & Partial \\
    Originality & \url{https://originality.ai/} & None \\
    GPTKit & \url{https://gptkit.ai/} & Partial \\
    CheckForAI & \url{https://checkforai.com/} & Partial  \\
    \bottomrule
  \end{tabular}
\end{table}

% Github repo: https://github.com/openai/gpt-2-output-dataset
\textbf{GPT-2 Output Detector}  \cite{Solaiman2019Release} is a LLM-generated text detector based on the RoBERTa large pretrained model \cite{robertapaper}. RoBERTa is a transformers model trained on a large corpus of raw English data. The GPT-2 Output Detector starts with the pre-trained ROBERTA-large model and trains a classifier for web data and the GPT-2 output dataset. The GPT-2 Detector returns the probability that an input text is real on GPT-2 text with accuracy of 88\% at 124 million parameters and 74\% at 1.5 billion parameters \cite{Solaiman2019Release}. The detector is limited to the first 510 tokens, although there are extensions that extend this limit \cite{GPT2Extension}.  \\

\textbf{GLTR} \cite{Sebastian2019GLTR} is a detector that applies statistical methods to detect GPT-2 text. The model is based on three simple tests: the probability of the word, the absolute rank of a word, and the entropy of the predicted distribution. This detector shows an interface where each word is highlighted along with a top-k class for that word. 

The GLTR detector does not provide quantifiable overall probability that a text is AI-generated. To make a fair comparison between GLTR and other detectors, we define a detector on top of GLTR to make probability predictions using the normal distribution. We compute an average $\mu$ and a standard deviation $\sigma$ over a sample dataset of 20 human and 20 ChatGPT submissions. The results were $\mu = 35.33$, and $s = 15.68$. We then used those results to normalize a prediction by computing the standard score of a data point $x$ using $\frac{x - \mu}{s}$. This score is sent as input to the sigmoid function to obtain a probability prediction.

\textbf{GPTZero} was the first detector \cite{GPTZeroTweet} to claim to detect ChatGPT data. The original version of the detector used two measures: perplexity and burstiness. Perplexity refers to a measurement of how well GPT-2 can predict the next word in the text. This appears similar to the way the GLTR detector works \cite{Sebastian2019GLTR}. The second measure is burstiness: the distribution of sentences. The idea is that humans tend to write with bursts of creativity and are more likely to have a mix of short and long sentence. The current version of GPTZero gives four classes of results. Table \ref{tab:gptzero} shows how different classes are interpreted as probability. GPTZero claims an 88\% accuracy for human text and 72\% accuracy for AI text for this detector \cite{GPTZeroEducation}.

\begin{table}
  \caption{GPTZero accuracy interpretation.}
  \label{tab:gptzero}
  \begin{tabular}{p{4.5cm} p{2.5cm}}
    \toprule Category & AI Probability \\ \midrule
Entirely written by human  & 0\% \\
    Likely entirely human, but some sentences have low perplexity & 20\%  \\
    May contain parts written by AI & 60\% \\
    Entirely written by AI &  100\%  \\
    \bottomrule
  \end{tabular}
\end{table}

\textbf{AI Text Classifier} is OpenAI's 2023 model fine tuned to distinguish between human-written and AI-generated text~\cite{AITextClassifier}. The model is trained on text generated from 34 models from 5 different organization. The model provides 5 different categories for the results based on the internal probabilities the model provides. Table \ref{tab:openai} shows how different classes are interpreted as probability. The interpretations are based on the final category, not the internal model. Usage of this classifier requires at least 1,000 characters.

\begin{table}
  \caption{AI Text Classifier interpretation.}
  \label{tab:openai}
  \begin{tabular}{p{3cm} p{2.5cm} p{2cm}}
    \toprule Category & Internal Probability & Interpretation \\ \midrule
    Very unlikely  & <0.1 & 0\% \\
    Unlikely  & 0.1 - 0.45 & 20\%  \\
    Unclear  & 0.45 - 0.9 & 50\% \\
    Possibly  &  0.9 - 0.98 & 75\%  \\
    Likely  &  >0.98 & 100\%  \\
    \bottomrule
  \end{tabular}
\end{table}

% TODO: Citation for each one?
\textbf{GPTKit} uses an ensemble of 6 other models, including DistilBERT \cite{DistilBERT}, GLTR, Perplexity, PPL, RoBERTa \cite{robertapaper}, and RoBERTa (base). The predictions of these models are used to form an overall probability that a text is LLM-generated. However, the exact weight used for each of the detectors is unclear. The detector claims an accuracy of 93\% based on testing on a dataset of 100K+ responses \cite{GPTKit}.

\textbf{CheckForAI} claims to combine the GPT-2 Output Detector along with custom models to help limit false readings \cite{CheckForAIDetector}. The detector also supports account sign up, history storage, and file uploads. The detector provides four classes to compute the probability of text, as shown in Table \ref{tab:checkforai}. This detector is currently limited to 2,500 characters.

\begin{table}
  \caption{CheckForAI accuracy interpretation.}
  \label{tab:checkforai}
  \begin{tabular}{p{4.5cm} p{2.5cm}}
    \toprule Category & AI Probability \\ \midrule
 Low Risk & 0\% \\
 Medium Risk    & 60\% \\
High Risk  &  80\%  \\
Very High Risk  &  100\%  \\
    \bottomrule
  \end{tabular}
\end{table}

\textbf{CopyLeaks} offers products for plagiarism and AI content detection targeted broadly for individuals, educators, and enterprises. The detector highlights paragraphs written by a human and by AI. CopyLeaks also claims detection across multiple languages, including Spanish (tested in this paper). CopyLeaks claims an accuracy of 99.12\% \cite{CopyLeaksDetector}. The detector is currently available publicly \cite{CopyLeaksDetector}.

\textbf{Originality.AI} is a detector targeted for content publishers. The detector is available through a commercial sign-up page \cite{OriginalityDetector} with a minimum fee \$20. We received research access for analysis of the detector. The detector comes with API access and a number of additional features for content creators. A self-proclaimed study by Originality on ChatGPT suggests that the detector has an accuracy of 98.65\%  \cite{Originalitystudy}.

We did not impose a systematic approach \cite{Kitchenham2009Systematic} to discover publicly available LLM-generated text detectors. Most of the detectors are recent and cannot be easily found on the internet or academic papers. A systematic approach might cover fewer results.

\subsection{Addressing the RQ: Effectiveness of LLM-generated text detectors}
A detector is only worthy of use if it is reasonably effective. We addressed the RQ by comparing detectors listed in Table \ref{tab:detectors} under three metrics: accuracy, false positives, and resilience. Instructors prefer to use detectors that are reasonably accurate, reporting a minimal number of false positives, and are resilient to disguises.

Accuracy refers to how effective the detectors are in identifying LLM-generated texts. We present all accuracy results using two measures of accuracy, as we have found that using only one measure may mislead about some aspect of the results.

The first method (averages) takes the average prediction each detector across a dataset. As discussed in the discovery section, each detector either provides a probability that a text is LLM-generated or a category that represents such a probability. We apply our category to AI conversion tables to obtain a probability for each detector. These probabilities are averaged for the final results. \\

The second method (thresholds) is calculated as the proportion of correctly-classified LLM-generated texts. These are measured as the number of texts that correctly receive above or below a 50\% score out of the total number of texts. This measure is strict, so a  prediction of 50\% is always considered to be incorrect.

% \textbf{Explain the additional data set, especially about the assessments}
% TODO

% setting the detectors
% \textbf{Explain here how each LLM-generated text detector was set to accept the input and how the first author obtained the `accuracy'.}

False positives are original submissions that are suspected by LLM-generated text detectors. Fewer false positives are preferred. For this metric, we collected student submissions before the release of ChatGPT (2019) and measured their degree of originality with the detectors. Any suspected submissions (originality degree less than 50\%) were expected to be false positives.

Resilience refers to how good LLM-generated text detectors are in removing disguises. Some students might disguise their LLM-generated texts to avoid getting caught. QuillBot \cite{Quillbot} is a  paraphrasing tool capable of paraphrasing text. The tool uses Artificial Intelligence to reword writing. We paraphrased 10 ChatGPT submissions through QuillBot and measured the results.

% \textbf{Things to discuss for this metric:
% \begin{itemize}
%     \item how to measure it
%     \item how to get the data sets (explain any tools used)
%     \item what are the disguises
% \end{itemize}}

It is worth noting that measuring effectiveness of LLM-generated text detectors is time consuming and labour intensive. Further, some detectors are not supported with API integration; the authors needed to manually copy and paste each test case. 

\subsection{Summarizing our experience using the LLM-generated text detectors}

We also report our experience in using the LLM-generated text detectors. Several aspects are considered: intuitiveness, clarity of documentation, extendability, variety of inputs, quality of reports, number of supported LLM-generated languages, and pricing.

\section{Results}
This section discusses our findings from addressing the research question and our experience using LLM-generated text detectors.

% This section analyzes and reports our findings, where Tables~\ref{tab:thresholdaccuracy} --~\ref{tab:quillbotweighted} presents the results of our English submissions and Tables~\ref{tab:spanishweighted} --~\ref{tab:spanishthreshold} presents the results of our submissions in Spanish.

\begin{table}
\caption{Overall accuracy of LLM-generated text detectors measured using thresholds. Sorted from best to worst.}
\label{tab:thresholdaccuracy}
\centering
\begin{tabular}{p{2.5cm} p{2cm} p{2cm}} \\ % p{1.5cm}}
\toprule
Detectors& Human Data & ChatGPT Data\\ % & Overall \\
\midrule
CopyLeaks & 99.12\% & 95.00\% \\ %& 97.06\% \\
GPT2 Detector & 98.25\% & 95.00\% \\ %& 96.62\% \\
CheckForAI & 98.25\% & 95.00\% \\ %& 96.62\% \\
GLTR & 82.46\% & 95.00\% \\ %& 88.73\% \\
GPTKit & 100.00\% & 75.00\% \\ %& 87.50\% \\
OriginalityAI & 93.86\% & 70.00\% \\ %& 81.93\% \\
AI Text Classifier & 94.74\% & 60.00\% \\ %& 77.37\% \\
GPTZero & 54.39\% & 45.00\% \\ %& 49.69\% \\
\bottomrule
\end{tabular}
\end{table}

\begin{table}
\caption{Accuracy of LLM-generated text detectors measured using weighted averages. Sorted from best to worst.}
\label{tab:weightedaccuracy}
\centering
\begin{tabular}{p{3cm} p{1.5cm} p{1.5cm}} % p{1.5cm}}
\toprule
Detectors & Human Data & ChatGPT Data \\ %& Overall \\
\midrule
CopyLeaks & 99.06\% & 94.14\% \\ %& 96.60\% \\
CheckForAI & 99.12\% & 94.00\% \\ %& 96.56\% \\
GPT2 Detector & 97.88\% & 94.70\% \\ %& 96.29\% \\
GPTKit & 95.13\% & 69.05\% \\ %& 82.09\% \\
AI Text Classifier & 96.49\% & 67.50\% \\ %& 82.00\% \\
OriginalityAI & 86.48\% & 66.77\% \\ %& 76.63\% \\
GLTR & 64.19\% & 67.48\% \\ %& 65.84\% \\
GPTZero & 73.95\% & 55.00\% \\ %& 64.47\% \\
\bottomrule
\end{tabular}
\end{table}

\begin{table}
\caption{False positive readings on LLM-generated text detectors. Sorted from best to worst.}
\label{tab:falsepositives}
\centering
\begin{tabular}{p{4cm} p{2.5cm}}
\toprule
Detectors & False Positives \\
\midrule
GPTKit & 0 \\
CopyLeaks & 1 \\
GPT2 Detector & 2 \\
CheckForAI & 2 \\
AI Text Classifier & 6 \\
OriginalityAI & 7 \\
GLTR & 20 \\
GPTZero & 52 \\
\bottomrule
\end{tabular}
\end{table}

\begin{table}
\caption{Resilience against Quillbot (accuracy). Sorted from highest to lowest accuracy.}
\label{tab:quillbot}
\centering
\begin{tabular}{p{2.5cm} p{1.5cm} p{1.5cm}}
\toprule
Detectors & Before & After \\
& Quillbot & Quillbot \\
\midrule
GLTR & 100.00\% & 100.00\% \\
GPT2 Detector & 100.00\% & 60.00\% \\
CopyLeaks & 100.00\% & 50.00\% \\
CheckForAI & 100.00\% & 40.00\% \\
OriginalityAI & 100.00\% & 40.00\% \\
GPTKit & 90.00\% & 30.00\% \\
AI Text Classifier & 60.00\% & 20.00\% \\
GPTZero & 70.00\% & 20.00\% \\
\bottomrule
\end{tabular}

\end{table}

\begin{table}
\caption{Resilience against Quillbot (weighted). Sorted from highest to lowest weight.}
\label{tab:quillbotweighted}
\centering
\begin{tabular}{p{2.5cm} p{1.5cm} p{1.5cm}}
\toprule
Detectors & Before & After \\
& Quillbot & Quillbot \\
\midrule
GLTR & 76.00\% & 62.57\% \\
GPTZero & 88.00\% & 62.00\% \\
AI Text Classifier & 67.50\% & 55.00\% \\
GPT2 Detector & 99.99\% & 54.95\% \\
CheckForAI & 100\% & 44.00\% \\
OriginalityAI & 90.24\% & 40.00\% \\
CopyLeaks & 100.00\% & 39.31\% \\
GPTKit & 82.60\% & 35.50\% \\
\bottomrule
\end{tabular}
\end{table}

\begin{table}
\centering
\caption{Accuracy of LLM-generated text detectors measured using weighted averages (using submissions in Spanish). Sorted from best to worst.}
\label{tab:spanishweighted}

\begin{tabular}{p{2.5cm} p{1.5cm} p{1.5cm} } %p{2cm}}
\toprule
Detectors & Human Data & ChatGPT Data \\ % & Overall \\
\midrule
AI Text Classifier & 22.50\% & 100.00\% \\ % & 61.25\% \\
OriginalityAI & 99.40\% & 19.89\% \\ %& 59.65\% \\
GLTR & 89.98\% & 20.70\% \\ %& 55.34\% \\
CopyLeaks & 1.31\% & 100.00\% \\ %& 50.66\% \\
GPT2 Detector & 99.98\% & 0.02\% \\ %& 50.00\% \\
CheckForAI & 100\% & 0.00\% \\ %& 50.00\% \\
GPTZero & 95.00\% & 0.00\% \\ %& 47.50\% \\
GPTKit & 57.60\% & 36.50\% \\ %& 47.05\% \\
\bottomrule
\end{tabular}
\end{table}

\begin{table}
\centering
\caption{Overall accuracy of LLM-generated text detectors measured using
thresholds (using submissions in Spanish). Sorted from best to worst.}
\label{tab:spanishthreshold}
\begin{tabular}{p{2cm} p{2cm} p{2cm} } %p{2cm}}
\toprule
Detectors & Human Data & ChatGPT Data \\ %& Overall \\
\midrule
OriginalityAI & 100.00\% & 20.00\% \\ %& 60.00\% \\
GPT2 Detector & 100.00\% & 0.00\% \\ %& 50.00\% \\
AI Text Classifier & 0.00\% & 100.00\% \\ %& 50.00\% \\
CopyLeaks & 0.00\% & 100.00\% \\ %& 50.00\% \\
CheckForAI & 100.00\% & 0.00\% \\ %& 50.00\% \\
GLTR & 100.00\% & 0.00\% \\ %& 50.00\% \\
GPTKit & 80.00\% & 20.00\% \\ %& 50.00\% \\
GPTZero & 90.00\% & 0.00\% \\ %& 45.00\% \\
\bottomrule
\end{tabular}
\end{table}

\subsection{Addressing the RQ: Effectiveness of LLM-generated Text Detector}

Table~\ref{tab:thresholdaccuracy} shows accuracy of each detector across human and ChatGPT data using the threshold method. The data shows CopyLeaks to be the most accurate LLM-generated text detector, with an accuracy of 97.06\%. CopyLeaks is followed by the GPT-2 Output Detector/CheckForAI (96.62\%), GLTR (88.73\%), GPTKit (87.50\%), OpenAI's Detector (77.37\%), and GPTZero (49.69\%). 

% The data in Table~\ref{tab:thresholdaccuracy} are normally distributed, verified by both the Shapiro-Wilk and Kolmogorov-Smirnov tests. Thus, no correction needed to be applied. Overall, from the t-test ($t = 1.67$, $p = 0.116$) we did not find significant differences in the accuracy of LLM-generated text detectors between human and chatgpt data.

Table \ref{tab:weightedaccuracy} shows the results using averages instead of thresholds. The results show CopyLeaks to provide the best probabilities (99.53\%), followed by CheckForAI (96.56\%), the GPT-2 Output Detector (96.29\%), GPTKit (82.09\%), OpenAI's Detector (82\%), OriginalityAI (76.63\%), GLTR (65.84\%), GPTZero (64.47\%).  

The data in Tables~\ref{tab:thresholdaccuracy} and~\ref{tab:weightedaccuracy} are both normally distributed, verified using the Shapiro-Wilk and Kolmogorov-Smirnov tests. Thus, no correction needed to be applied. Overall, from the t-tests (Table~\ref{tab:thresholdaccuracy}: $t = 1.67$ and $p = 0.116$, Table~\ref{tab:weightedaccuracy}: $t = 1.154$, $p = 0.268$, both with 14 degrees of freedom) we did not find significant differences in the accuracy of LLM-generated text detectors between human and ChatGPT data.

Table \ref{tab:falsepositives} shows the false positive results on the human data from the databases and network assignments. GPTKit is the only detector that managed to achieve no false positives across the entire set of human submissions. This is followed by CopyLeaks (1), the GPT-2 Output Detector/CheckForAI (2), OpenAI's detector (6), OriginalityAI (7), GLTR (20), and finally GPTZero (52).

\begin{figure}
    \caption{Report introduction with 37\% AI Probability on GPTKit.}
    \label{fig:gptkit37}
    \centering
    \includegraphics[width=0.5\textwidth]{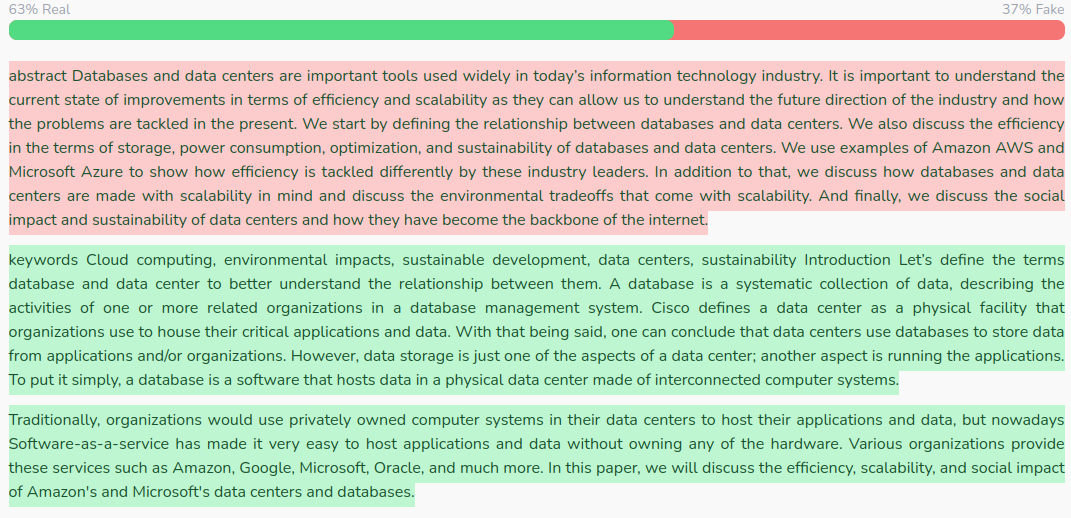}
\end{figure}

\begin{figure}
\caption{Truncated introduction with 54\% AI Probability on GPTKit.}
    \label{fig:gptkit54}
    \centering
    \includegraphics[width=0.5\textwidth]{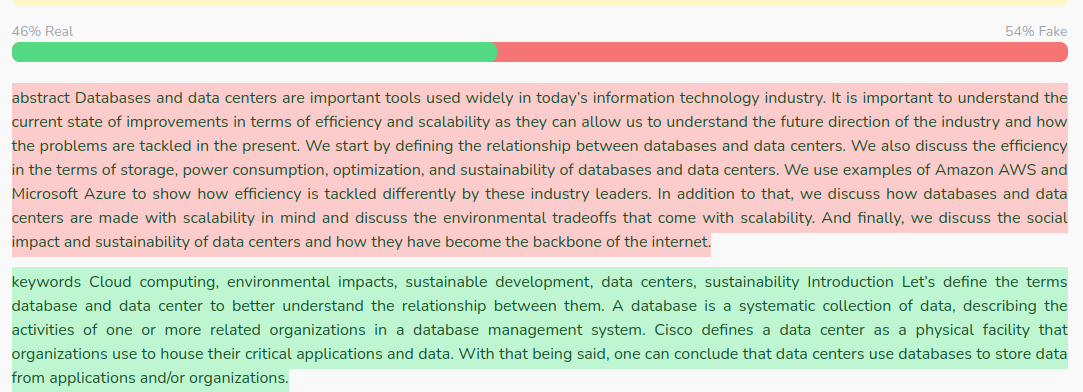}
\end{figure}

A further investigation of GPTKit, which appears to be the the best detector for avoiding false positives, shows that this detector is still prone to false positives. While none of our original test samples appeared more than 50\% fake, we found that some submissions score up to 37\% fake from GPTKit. In some cases, removing the last paragraph(s) from these submissions led to a false positive. Figures \ref{fig:gptkit37} and \ref{fig:gptkit54} show such a case. We note that in this case the output of GPTKit also shows that the detector merged separate paragraphs into a single one. This unexpected merge may contribute to the problem.

% Note: We could do a weighted version of this one. 
Table \ref{tab:quillbot} shows results of 10 ChatGPT papers before and after the Quillbot paraphraser. The results are measured using overall accuracy. The GLTR detector was the most resilient, with none of the predictions changing. It is worth noting that the overall weighted result of GLTR also decreased by 10\%, although the change did not effect the accuracy. In contrast, the rest of the detector saw a significant drop following the transformation of Quillbot.

\begin{figure}
\caption{ChatGPT report plugged to Originality AI before Quilltbot .}
    \label{fig:originalitybefore}
    \centering
    \includegraphics[width=0.5\textwidth]{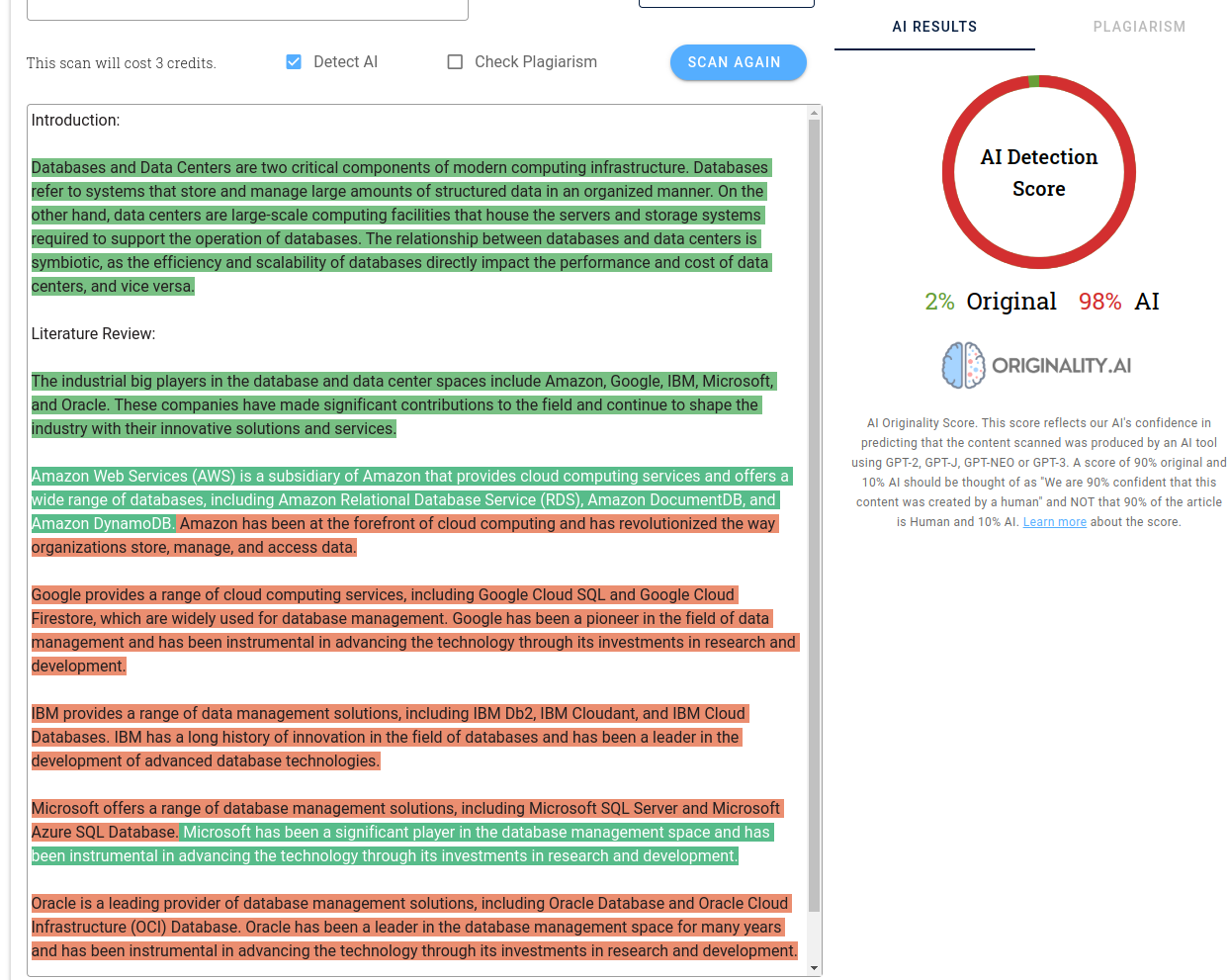}
\end{figure}

\begin{figure}
\caption{ChatGPT report plugged to Originality AI after Quilltbot .}
    \label{fig:originalityafter}
    \centering
    \includegraphics[width=0.5\textwidth]{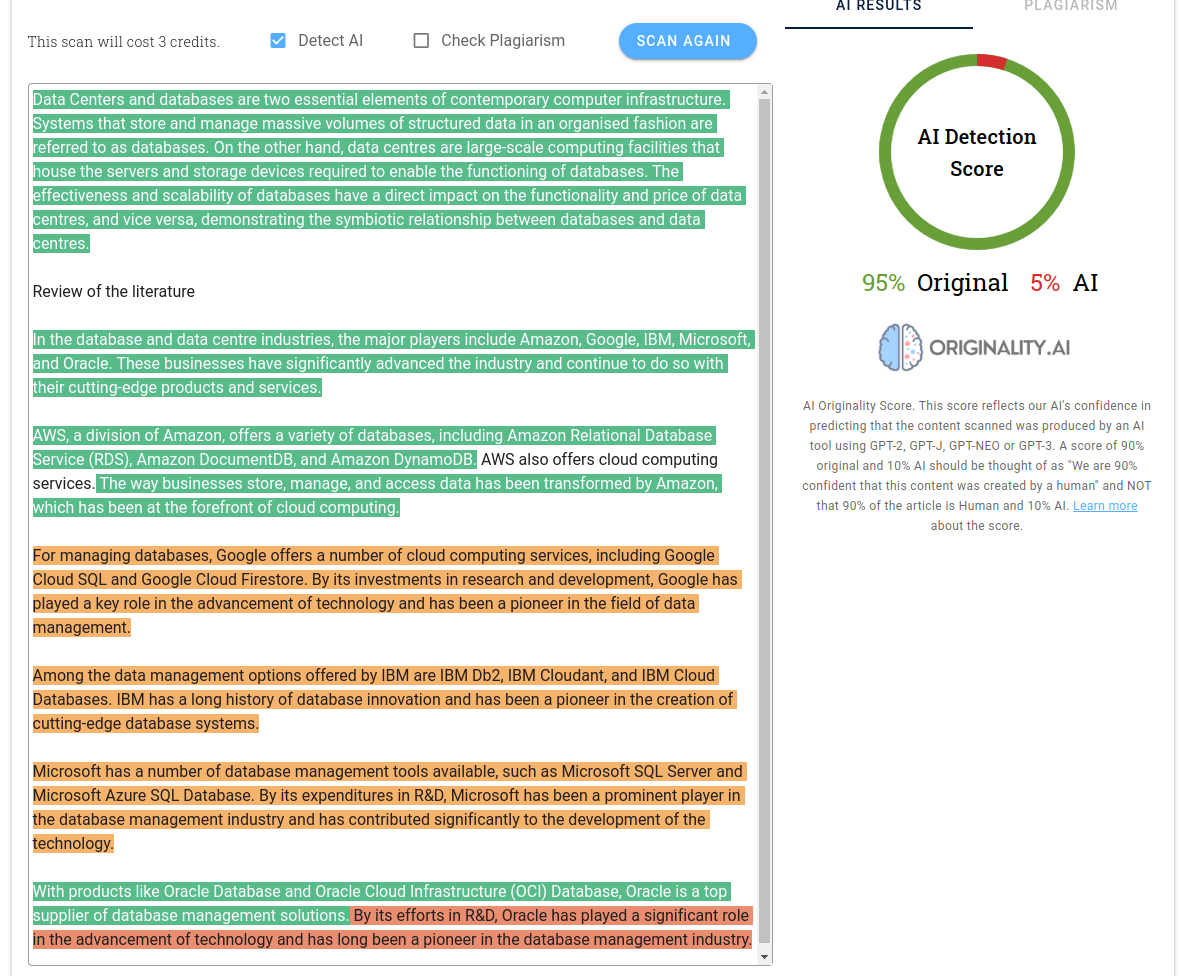}
\end{figure}

Figures \ref{fig:originalitybefore} and \ref{fig:originalityafter} show an example of a ChatGPT data point that went from 98\% before Quillbot to 5\% after Quillbot on Originality. 
% TODO: More thoughts bout this? 

Tables \ref{tab:spanishweighted} and \ref{tab:spanishthreshold} show results from the capstone course data, written using Spanish. We found that CopyLeaks and the AI Text Classifier tend always output fake predictions on AI data. In contrast, the GPT-2 Output Detector, GPTZero, CheckForAI, GLTR, GPTKit, and Originality tend to output human predictions.

The data in Tables~\ref{tab:spanishweighted} and~\ref{tab:spanishthreshold} are both normally distributed, verified using the Shapiro-Wilk and Kolmogorov-Smirnov tests. Thus, no correction needed to be applied. Overall, from the t-tests (Table~\ref{tab:spanishweighted}: $t = 1.766$ and $p = 0.099$, Table~\ref{tab:spanishthreshold}: $t = 1.862$, $p = 0.084$, both with 14 degrees of freedom) we did not find significant differences in the accuracy of LLM-generated text detectors between human (Spanish text) and ChatGPT (Spanish text) data.

The GLTR detector shows an interesting mild success with Spanish data. The average top-k score on human data was $104$, while the average top-$k$ score on ChatGPT data was $85$. When we changed the implementation of GLTR to set a mean of a $94.5$ top-k score, GLTR managed to achieve the highest accuracy of $65\%$ on Spanish text.

% \subsection{Addressing RQ2: Usability of LLM-generated Text Detectors}
\subsection{Our experience using the LLM-generated text detectors}

% We generally discuss the usability of LLM-generated text detectors based on a number of usability aspects.

% \subsubsection{Intuitiveness}
Generally, many LLM-generated text detectors are intuitive to use. Similar with many online similarity detectors for identifying text plagiarism \cite{Blanchard2022Stop}. They have a web-based interface where a user can paste the text they want to check its originality. GPTZero and CheckForAI allow their users to upload a document instead.

% \subsubsection{Clarity of Documentation}
While there are a number of LLM-generated text detectors, only two of them have their technical reports publicly available (GPT-2 Output Detector \cite{Solaiman2019Release} and GLTR \cite{Sebastian2019GLTR}). This is possibly due to at least two reasons. First, technical reports might be misused by some individuals to trick the detectors. Second, some detectors are commercial.

% \subsubsection{Extendability}
Most LLM-generated text detectors do not facilitate API integration. GPTZero, GPTKit, OriginalityAI, CopyLeaks provide such a feature with a fee. Without API integration, it is challenging to integrate the detectors to existing teaching environments, especially learning management system. LLM-generated text detectors are unlikely to be independently used as the task is labor intensive.

As many of the detectors are commercial, their code is not publicly available. This might complicate instructors to further develop the detectors to fit their particular needs. The only open source detectors are the GPT-2 Output Detection and GLTR.

% \subsubsection{Variety of inputs}

The detectors are also limited in the input formats they support. Most of them only allow raw text pasted in a form, making them difficult to automate. The PDF parsers that we attempted to use often parsed in an incorrect order and had a tendency to include unwanted characters. We had to write custom scripts to parse the text in a format that translates all information to text. 

% \subsubsection{Quality of Reports}
Detection results are challenging to interpret. Detectors attempt to combat this problem by highlighting content that is more likely to be AI-generated. Table~\ref{tab:highlighting} shows the highlighting support each detector provides. Highlighting is provided on either a paragraph, sentence, or a word basis.

\begin{table}
\caption{Highlighting support per detector.}
\label{tab:highlighting}
\centering
\begin{tabular}{p{3cm} p{5cm} }
\toprule
Detector & Highlighting support \\
\midrule
GPT2 Detector & None \\
AI Text Classifier & None  \\
CopyLeaks & Paragraphs \\
CheckForAI & Sentences \\
GPTZero & Sentences \\
GLTR & Words \\
GPTKit & Paragraphs \\
OriginalityAI & Sentences  \\
\bottomrule
\end{tabular}

\end{table}

\begin{figure}
    \caption{A false positive using GPTZero.}
    \label{fig:gptzeromay}
    \centering
    \includegraphics[width=0.5\textwidth]{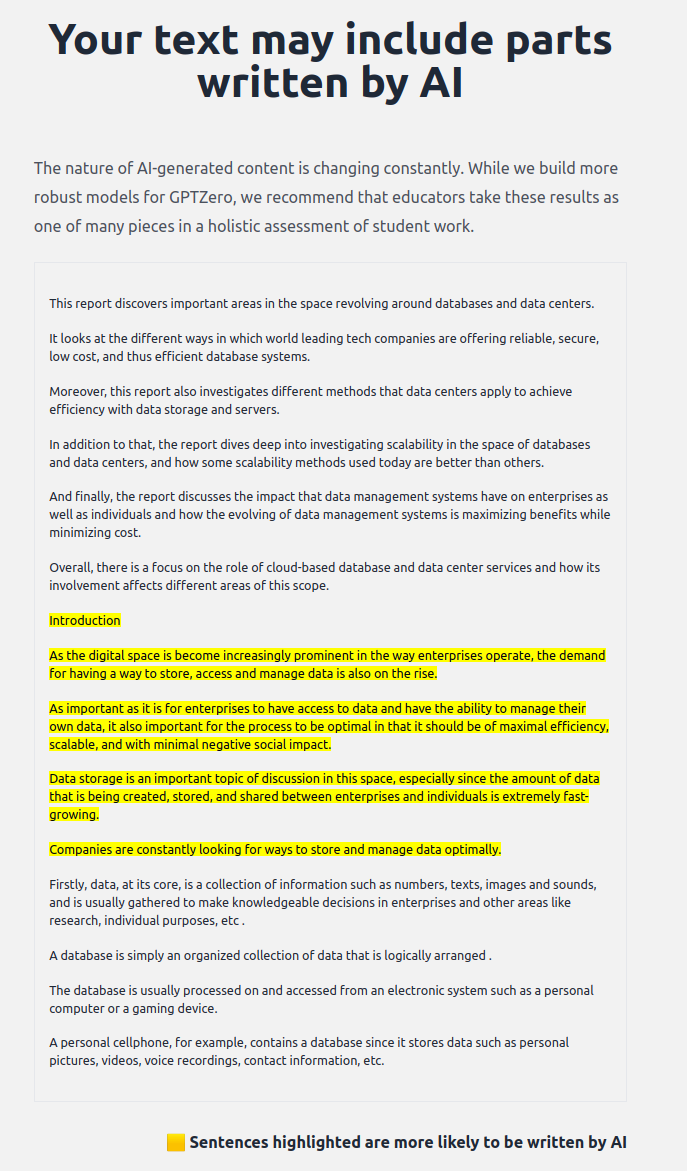}
\end{figure}

While highlighting does seem to mitigate some barriers, we found that the highlighting feature can still be misleading. This was particularly evident in GPTZero, which highlighted 52 human submissions as either possibly or entirely AI-generated. Figure \ref{fig:gptzeromay} shows a sample human report where some sentences were highlighted as more likely to be written by AI. It is unclear what makes the highlighted text more likely be written by AI than the other sentences.

In terms of output quality, it seems like the detectors are limited in their ability to export results. Nevertheless, some detectors were more effective than others. We provided screenshots of GPTKit, GPTZero, and Originality in this report since they provided more detailed results and it was easier to screenshot the results along with the text in contrast to the other detectors. It was more challenging to show full results of other detectors as they did not allow side-by-side results. 

% \subsubsection{Number of Supported LLM-generated Languages}

Most LLM-generated text detectors only support English as the language of LLM-generated text. While one can still send text in other languages, the results do not appear meaningful as we previously showed. 

% \subsubsection{Pricing}
As many LLM-generated text detectors are commercial and they are relatively new, there appear to mostly individual pricing options. GPTZero  CopyLeaks, for instance, have business pricing. GPTZero currently has a subscription plan for business users for $\$19.99$USD per month.

These detectors might be far less useful for instructors living in countries with weak currency; the pricing options are only available in USD.

\section{Discussion}

The current state of LLM-generated text detectors suggests that they are not yet ready to be trusted blindly for academic integrity purposes or as reliable plagiarism detectors such as Turnitin, MOSS, or JPlag. Our study demonstrates that detectors under-perform compared to the GPT-2 Output Detector and GLTR, which are older and freely available detectors from 2019. 

%From thoroughly testing these detectors up until April 2023, 
At first glance, it appears that LLM-generated text detectors are fairly accurate with human data being correctly detected $\sim89.59\%$\footnote{this percentage is the average accuracy for human data using Tables~\ref{tab:thresholdaccuracy} and~\ref{tab:weightedaccuracy}.} while the average accuracy for ChatGPT-generated data is substantially lower; $\sim77.42\%$\footnote{this percentage is the average accuracy for ChatGPT-generated data using Tables~\ref{tab:thresholdaccuracy} and~\ref{tab:weightedaccuracy}.}. Upon deeper inspection, it is apparent that the number of potential false positives can lead to a wide array of issues, especially if being trusted for plagiarism detection at educational institutions. 

Delving further, when a paraphraser (in this case, QuillBot) is utilized the average accuracy is slightly reduced for human data  
$\sim89.02\%$\footnote{this percentage is the average accuracy for human data using Tables~\ref{tab:quillbot} and~\ref{tab:quillbotweighted}.} but this substantially reduces the accuracy of ChatGPT-generated data $\sim49.17\%$\footnote{this percentage is the average accuracy for ChatGPT-generated data using Tables~\ref{tab:quillbot} and~\ref{tab:quillbotweighted}.}. This means that in more than half of all cases, ChatGPT-generated data cannot correctly be identified by these detectors. Though, some detectors perform better than others (e.g., GLTR), it is still a serious concern for users of these detectors. 

Additionally, once non-English languages are introduced, these detectors are easily exacerbated. We investigate submissions made in Spanish and see that the average accuracy for human data lowers to an average of $\sim70.99\%$ \footnote{this percentage is the average accuracy for human data using Tables~\ref{tab:spanishweighted} and~\ref{tab:spanishthreshold}.}, and ChatGPT-generated data reduces to an abysmal $\sim17.50\%$\footnote{this percentage is the average accuracy for ChatGPT-generated data using Tables~\ref{tab:spanishweighted} and~\ref{tab:spanishthreshold}.}. Though only Spanish was investigated, it introduces the need for additional research into alternative languages (non-English).

Presently, all LLM-generated text detectors struggle with languages other than English, code, and special symbols, resulting in fairly inaccurate results. As a point of clarity, it would be ideal for these detectors to explicitly state their limitations and aim to produce human predictions in such cases.

% LLM-generated text detectors are vulnerable to minor manipulations. We found that using a paraphraser like Quillbot reduced the AI-generated probabilities by an average of 50\%. Additional human interventions and mixed content can further deceive detectors, leading them to classify AI content as human-generated.

% All LLM-generated text detectors struggle with languages other than English, code, and special symbols, resulting in inaccurate results. To improve performance, detectors should explicitly state their limitations and aim to produce human predictions in such cases. Moreover, educators must be proficient in running LLM-generated text detectors and vigilant for any parsing issues.

In terms of usability, LLM-generated text detectors need some improvements. Although they are intuitive to use and generate acceptable reports, many of them are not well documented at a technical level, some do not have APIs making them more difficult to integrate into local and larger systems (e.g., Learning Management Systems), and the support of these detectors is limited. Furthermore, some of these detectors require processing fees.   

From our results, LLM-generated text detectors appear to lack in understandability. We are aware that all of these detectors leverage similar large language models for detection purposes. However, they might differ in terms of their technical implementation, parameters, pre-trained data, etc. These are unlikely to be revealed since most of the detectors are for commercial-use and, thus, proprietary. While some detectors highlight sentences that are more likely to be AI-generated (Table~\ref{tab:highlighting}), the results produced by the detectors  are not clear enough for users of these detectors.

% LLM-generated text detector creators must acknowledge these limitations and prioritize human predictions in unfamiliar domains. Presently, all LLM-generated text detectors struggle with languages other than English, code, and special symbols, resulting in inaccurate results. To improve performance, detectors should explicitly state their limitations and aim to produce human predictions in such cases. Moreover, educators must be proficient in running LLM-generated text detectors and vigilant for any parsing issues.

% Another concern is that LLM-generated text detectors are vulnerable to minor manipulations. We found that using a paraphraser like Quillbot reduced the AI-generated probabilities by an average of 50\%. Additional human interventions and mixed content can further deceive detectors, leading them to classify AI content as human-generated.

% In terms of usability, LLM-generated text detectors need some improvements. Although they are intuitive to use and generate acceptable reports, many of them are not well documented at technical level, are not extendable, and support limited types of inputs and LLM-generated languages. Further, some of them require processing fee.   

% As the detectors are not quite effective and have a number of limitations, improvements are needed before the detectors can be used for any educational or commercial applications.   

\section{Threats to Validity}
Our study has several threats to validity:
\begin{itemize}
    \item The findings of the study reflect detector results that are accurate as of April 2023. The detectors are volatile, and owners of these detectors could update their models. Results could change based on updates to LLM-generated text detectors.
    \item Accuracy, false positives, and resilience were arguably sufficient to represent effectiveness. However, additional findings can be obtained by considering other effectiveness metrics. 
    \item The data sets were obtained from two institutions; one uses English as the operational language while another uses Spanish. This means that the findings might not be generalizable to other institutions, especially those with different operational languages.
    \item While we believe that the data sets are sufficient to support our findings, we acknowledge that more data sets can strengthen the findings.
    % \item To assess practicality of LLM-generated text detectors, our practical factors were defined by us as we have some experience in maintaining academic integrity. However, the list was not exhaustive and more factors might be useful for consideration.
\end{itemize}

\section{Conclusion}
This paper examines eight LLM-generated text detectors on the basis of effectiveness. The paper shows that while detectors manage to achieve a reasonable accuracy, they are still prone to flaws and can be challenging to interpret by the human eye. Ultimately, LLM-generated text detectors, while not yet reliable for academic integrity or plagiarism detection, show relatively accurate results for human-generated data compared to ChatGPT-generated data. However, false positives are a significant concern, especially when used for plagiarism detection in educational institutions. When a paraphrasing tool like QuillBot is employed, the accuracy decreases for both human and ChatGPT-generated data. Additionally, the detectors struggle with non-English languages, resulting in even lower accuracy. It is crucial for these detectors to acknowledge their limitations and aim for improved performance in various language contexts.

% In summary, some key findings during these research is as follows:

% \begin{itemize}
%     \item When tested on pure English text, most detectors achieved an accuracy of 80\% or above, with CopyLeaks scoring the highest (97\%) and GPTZero scoring the lowest (49.69\%). However, accuracy is just one metric and it's important to consider a holistic view.   
%     \item All detectors were prone to the problem of false positives. While GPTKit is the best detector to avoid false positives, we found human examples that were also misclassified by GPTKit. 
%     \item The use of paraphraser tools severely affects the ability of LLM-generated text detectors, making it very easy for students to use a LLM and then a tool (such as Quillbot) to make the text seem human generates. GLTR seems to be the most resilient LLM-generated text detector on the market.
%     \item The current LLM-generated text detectors are not trained on non-English languages, so using these tools for non-English text (like Spanish) is pointless. 
% \end{itemize}

% Ultimately, it is clear that these LLM-generated text detectors should not be used in a commercial or education setting as they are highly unreliable and it is extremely easy to fool them.

\subsection{Future Work} 
Future detectors could attempt to incorporate a combination of metrics along with their accuracy for AI detectors. A combination of many factors along with the accuracy and false positive rates may give educators better insights into the predictions. This could include text-based features such as burstiness and repetition as well as AI-learned features such as probabilities. These detectors could further be fine-tuned for specific domains to improve their reliability.

Additionally, there is a fundamental need to have accurate and understandable LLM-generated text detectors available for educators to combat against the rising concern of academic integrity due to these publicly available LLMs. It is also important for the researchers to contact the creators of these detectors to better understand the related issues and needs of the end users, but also to facilitate a deeper conversation about the functionality and correctness of their instruments.

Finally, there is an apparent need to investigate the use of non-English languages using these detectors as large language models, like the one(s) used by ChatGPT, can produce content in languages other than English.

% \section{Acknowledgements}
% This research was supported by the University of Toronto Mississauga’s Office of the Vice-Principal, Research. Additionally, we would like to thank \href{mailto:ccperez@espol.edu.ec}{Cinthia Pérez}, from ESPOL, for aiding us in sourcing some of the data. We would also like to thank \href{https://harsh-kumar.com}{Harsh Kumar} and \href{https://www.naazsibia.com}{Naaz Sibia} for proofreading this work.